\documentclass[10pt,conference]{IEEEtran}
\IEEEoverridecommandlockouts
\usepackage{subfigure}
\usepackage{cite}
\usepackage{amsmath,amssymb,amsfonts}
\usepackage{algorithmic}
\usepackage{algorithm}
\usepackage{graphicx}
\usepackage{textcomp}
\usepackage{xcolor}
\usepackage{url}
\usepackage{bm}
\begin{document}

\title{Loss-Resilient Wireless Video Token Communication over Block Fading Channels}
\author{Bingyan Xie$^1$, Yongjeong Oh$^2$, Zihan Chen$^2$, Jihong Park$^2$, Yongpeng Wu$^1$, Wenjun Zhang$^1$\\
	$^1$Department of Electronic Engineering, Shanghai Jiao Tong University, Shanghai 200240, China\\
	$^2$ISTD Pillar, Singapore University of Technology and Design, 8 Somapah Rd, Singapore 487372\\
	Email:$\{$bingyanxie, yongpeng.wu, zhangwenjun$\}$@sjtu.edu.cn,\\ $\{$yongjeong\_oh, jihong\_park$\}$@sutd.edu.sg, zihan\_chen@mymail.sutd.edu.sg	
}

\maketitle

\begin{abstract}
	Video token communication represents video content as discrete tokens that differ in their importance to reconstruction and exhibit temporal dependencies. When these tokens are packetized for wireless transmission, block fading can cause multiple important or correlated tokens to be lost together, severely degrading video reconstruction.
	To address this issue, we propose a loss-resilient wireless video token
	communication (WVTC) framework. WVTC evaluates token importance from
	the intrinsic predictive structure of video tokens, assigning high
	priority to structural I-tokens and measuring P-token importance by
	temporal neighborhood novelty. A shuffled mixed I/P-token packetization
	scheme disperses structural anchors and correlated temporal regions
	across packets. Using only current block channel state information, an
	online scheduler jointly considers packet importance density,
	MCS-dependent decoding reliability, block capacity, and importance
	concentration when allocating packets to fading blocks. At the receiver,
	a fine-tuned detokenizer reconstructs missing content from surviving
	tokens without retransmission. Numerical results demonstrate improved
	perceptual quality and more graceful degradation under increasing packet
	error rates.
\end{abstract}


\section{Introduction}

Driven by recent advances in generative models, token communication
(TokenCom) has emerged as a promising communication approach that
leverages discrete tokens as the basic units for representing and
conveying source content~\cite{tokencom}. These tokens capture compact
and context-aware representations, enabling communication systems to
selectively process and protect information at the token level.
Moreover, their contextual dependencies can be exploited at the
receiver to recover missing information, making TokenCom particularly
attractive for error-prone wireless transmission.

Early TokenCom studies primarily focused on text and image modalities and
their cross-modal combinations~\cite{tokencom,todma}. However, despite
its potential for efficient video transmission, video TokenCom remains
relatively underexplored because strong spatial and temporal
dependencies among video tokens complicate the evaluation of token
importance and the impact of token losses on video reconstruction.
As an early work on video TokenCom, Morphe employs adaptive token dropping
based on token similarity and available bandwidth to reduce transmission
overhead, while using a vision foundation model to reconstruct the video
from retained tokens~\cite{morphe}. Video TokenCom instead utilizes
textual descriptions to identify important video tokens and allocates
more bits and stronger channel protection to them~\cite{Vidtokencom}.
However, these studies mainly focus on reducing transmission overhead
or protecting important tokens at the token level, without considering
how video tokens are packetized and transmitted in practical
packet-based wireless systems. As a result, conventional packetization
that ignores token importance and temporal correlation may group
critical or strongly related tokens into the same packet, making a
single packet loss disproportionately harmful to video reconstruction.
This problem becomes more severe over block fading channels, where
multiple packets transmitted within the same fading block can be
simultaneously affected by a deep fade.

Conventional wireless systems mitigate packet losses using forward
error correction (FEC), retransmission, and interleaving. However, FEC
requires additional redundancy and retransmission incurs extra latency,
while conventional interleaving disperses data without considering the
different roles and temporal dependencies of video tokens. Therefore,
directly applying these techniques to video TokenCom is inefficient,
motivating a packet-level transmission design that explicitly accounts
for the structural characteristics of video tokens. Specifically, the
first token group contains structural intra-coded (I)-tokens, while
subsequent inter-coded (P)-token groups progressively represent temporal
changes. Since I-tokens provide structural information and P-tokens
exhibit temporal correlations, concentrating important or correlated
tokens in the same packet can cause a single packet erasure to remove
multiple critical tokens at once. This risk becomes more severe over
block fading channels, where packets within the same fading block
experience the same channel condition and can be lost together during
a deep fade. Therefore, robust video TokenCom requires the joint design
of token-to-packet mapping and packet-to-block allocation to disperse
important and correlated tokens across packets and fading blocks while
adapting their transmission to current channel conditions.

\begin{figure*}[htbp]
	\centering
	\includegraphics[width=5.4in]{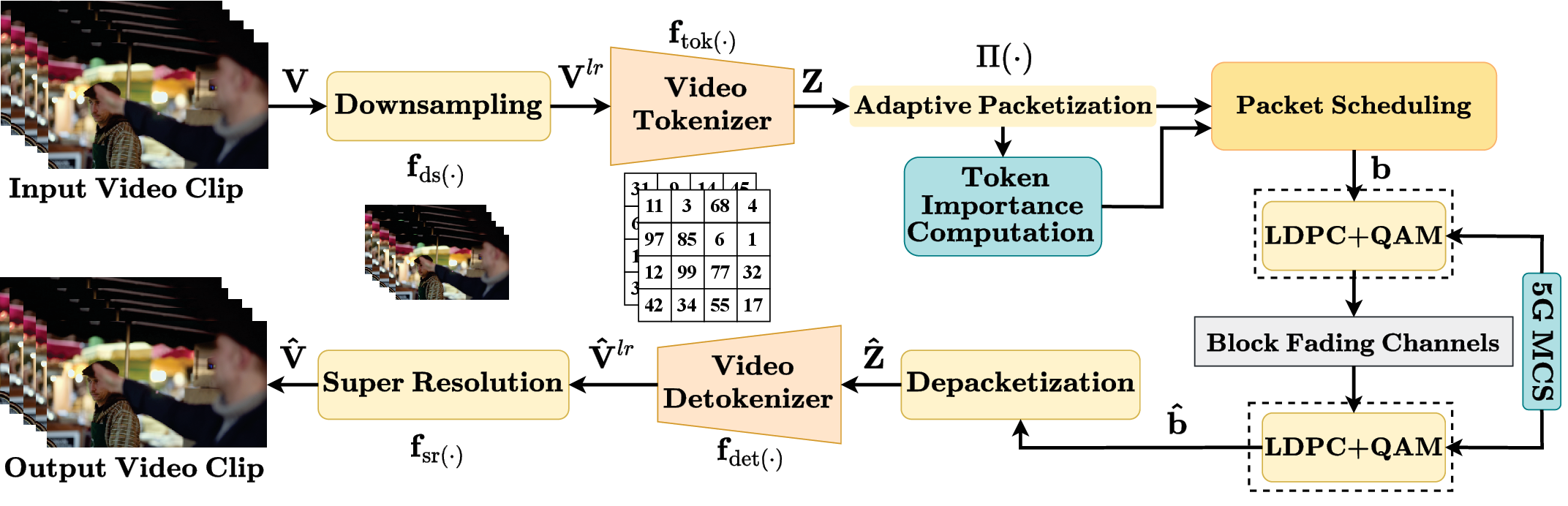}
	\vspace{-12pt}
	\caption{System model of the proposed loss-resilient wireless video
		token communication framework over block fading channels.}
	\vspace{-12pt}
	\label{fig_1}
\end{figure*}

In this paper, we propose a loss-resilient wireless video token
communication (WVTC) framework for block fading channels. WVTC first
evaluates token importance from the structural and temporal
characteristics of video tokens. It then disperses important and
correlated I/P-tokens across packets and allocates the resulting packets
online according to their importance and current channel state
information (CSI). At the receiver, an erasure-aware detokenizer
reconstructs missing video content from the successfully received
tokens without retransmission. The main contributions are summarized
as follows
\begin{itemize}
	\item We develop a video-token importance metric that distinguishes the
	roles of I- and P-tokens in reconstruction. I-tokens are assigned high
	priority as structural anchors, while P-token importance is measured by
	local representation changes across adjacent token groups, without
	requiring textual prompts or an additional vision-language model.
	
	\item We propose an importance-aware packetization scheme that disperses
	structural I-tokens and temporally correlated P-tokens across packets,
	reducing concentrated information loss under packet erasures without
	explicit token-position signaling.
	
	\item We design an online CSI-aware packet allocation algorithm for
	block fading channels. Using only current-block CSI, it jointly considers
	packet importance, decoding reliability, block capacity, and importance
	concentration, reducing the risk that multiple important packets are lost
	under the same deep fade without retransmission.
\end{itemize}

\textit{Notations:} 
$\mathbb R$ denotes the real number set, and $\mathcal K$ the discrete
token alphabet. $\mathcal{CN}(\boldsymbol{\mu},\boldsymbol{\Sigma})$
and $\mathcal U(a,b)$ denote the complex Gaussian and uniform
distributions, respectively. $\mathbf I$ and $\varnothing$ denote the
identity matrix and empty set. For a scalar $a$, $|a|$ denotes its
magnitude, while $|\mathcal A|$ denotes the cardinality of set
$\mathcal A$. $\mathcal A\setminus\mathcal B$, $\lceil x\rceil$, and
$(\cdot)^{\mathsf T}$ denote set difference, ceiling, and transpose,
respectively.

\section{System Model}

As shown in Fig.~\ref{fig_1}, we consider point-to-point wireless video
token communication over block fading channels.

\subsection{Video Tokenization and Packetization}

Let $\mathbf V=\{\mathbf x_t\}_{t=1}^{T}$ denote an input video clip
with $T$ frames, where
$\mathbf x_t\in\mathbb R^{3\times H_0\times W_0}$ is the $t$-th
frame, and $H_0$ and $W_0$ are the frame height and width. The clip is
first downsampled as
$\mathbf V^{\mathrm{lr}}=f_{\mathrm{ds}}(\mathbf V)$, where
$f_{\mathrm{ds}}(\cdot)$ denotes the downsampling operation. The
tokenizer $f_{\mathrm{tok}}(\cdot)$ then maps the low-resolution clip
into
$\mathbf Z=f_{\mathrm{tok}}(\mathbf V^{\mathrm{lr}})
\in\mathcal K^{G\times H_z\times W_z}$, where $\mathcal K$ is the
discrete token alphabet, $G$ is the token group number, and
$H_z\times W_z$ is the group spatial size. Let
$z_{g,u,v}=[\mathbf Z]_{g,u,v}\in\mathcal K$ denote the token index at
position $(g,u,v)$. The first group contains structural I-tokens, while
the remaining $G-1$ groups contain temporal P-tokens that progressively
represent changes relative to the preceding groups.

The packetization function $\Pi(\cdot)$ partitions the token indices
into $N_p$ packets as
$\{\mathcal P_n\}_{n=1}^{N_p}=\Pi(\mathbf Z)$, where $\mathcal P_n$
contains the token indices assigned to packet $n$ according to the
packetization mapping introduced in Sec.~III-B. Since the packetization
rule and shuffle seed are shared at both ends, the original token
positions can be recovered from the packet ID without explicit position
signaling.

After attaching the packet ID and cyclic redundancy check (CRC), the
resulting packet is converted into a bitstream
$\mathbf b_n\in\{0,1\}^{L_n}$, where $L_n$ denotes the total bit
length, including the token payload and signaling overhead. To control
the transmission rate, only a subset of the generated packets is
selected for transmission based on their importance. We denote the
indices of these selected packets by
$\mathcal Q_1\subseteq\{1,\ldots,N_p\}$.

\subsection{Block Fading Transmission and Packet Scheduling}

The wireless channel follows a block fading model with $B$ fading
blocks, where the channel coefficient remains constant within each
block and varies independently across blocks. For block
$b\in\{1,\ldots,B\}$, the channel coefficient is modeled as
$h_b\sim\mathcal{CN}(0,1)$.

At the beginning of block $b$, the transmitter obtains the current
CSI through channel estimation and
feedback. Let $\gamma_b$ denote the corresponding signal-to-noise ratio
(SNR), and $\mu(\cdot)$ the predefined SNR-MCS lookup table. The
modulation and coding scheme (MCS) index is selected as
$m_b=\mu(\gamma_b)$. All packets transmitted in the same block use the
same MCS, and no future CSI is assumed to be available.

Let $\mathcal Q_b$ denote the retained packets that remain unscheduled
before block $b$. Under MCS $m_b$, let $C_b$ denote the source-bit
capacity of block $b$, determined by the available channel uses and the
coding rate and modulation order associated with $m_b$. Let
$\mathcal S_b\subseteq\mathcal Q_b$ denote the packet-index set
scheduled in block $b$, satisfying
\begin{equation}
	\mathcal S_b\subseteq\mathcal Q_b,\qquad
	\sum_{n\in\mathcal S_b}L_n\leq C_b,\qquad
	\mathcal Q_{b+1}
	=
	\mathcal Q_b\setminus\mathcal S_b.
	\label{eq:scheduling_constraint}
\end{equation}

The process continues until $\mathcal Q_{B+1}=\varnothing$, indicating
that all retained packets have been scheduled.

For each scheduled packet $\mathcal P_n$, $n\in\mathcal S_b$, its
bitstream $\mathbf b_n$ is channel-coded and modulated according to
$m_b$, producing the transmit symbol vector $\mathbf s_{n,b}$. The
received signal is
\begin{equation}
	\mathbf y_{n,b}
	=
	h_b\mathbf s_{n,b}
	+
	\mathbf w_{n,b},
	\label{eq:received_signal}
\end{equation}
where
$\mathbf w_{n,b}\sim
\mathcal{CN}(\mathbf 0,\sigma_b^2\mathbf I)$
denotes additive white Gaussian noise (AWGN) with noise variance
$\sigma_b^2$. Accordingly, the instantaneous SNR of block $b$ is
\begin{equation}
	\gamma_b
	=
	\frac{|h_b|^2P_s}{\sigma_b^2},
	\label{eq:instantaneous_snr}
\end{equation}
where $P_s$ denotes the average transmit-symbol power.

After equalization, demodulation, channel decoding, and CRC checking,
a packet is either successfully recovered or treated as erased. Its
decoding reliability is characterized by
\begin{equation}
	R_{b,n}
	=
	1-\operatorname{BLER}(\gamma_b,m_b,L_n),
	\label{eq:packet_reliability}
\end{equation}
where $\operatorname{BLER}(\gamma_b,m_b,L_n)$ denotes the block error
rate, which depends on the SNR, MCS index, and packet length.

\subsection{Receiver Reconstruction}

The receiver collects all successfully decoded packets and restores
their token indices to the original token-grid positions according to
the shared packetization rule and packet IDs. Let
$\widehat{\mathbf Z}$ denote the resulting incomplete token tensor,
which has the same dimensions as $\mathbf Z$. Token indices carried by
successfully decoded packets are restored to their corresponding
positions, whereas positions belonging to erased packets are filled
with zeros. No explicit reception mask is provided to the reconstruction
network.

The detokenizer reconstructs the low-resolution
video directly from the zero-filled token tensor as
$\widehat{\mathbf V}^{\mathrm{lr}}
=f_{\mathrm{det}}(\widehat{\mathbf Z})$. By exploiting the surviving
I- and P-token context, the fine-tuned detokenizer performs implicit
spatiotemporal inpainting of the missing content. Thus, no separate
token-prediction module or packet retransmission is required. The
super-resolution network $f_{\mathrm{sr}}(\cdot)$ subsequently
produces the final reconstructed video clip as
$\widehat{\mathbf V}
=f_{\mathrm{sr}}(\widehat{\mathbf V}^{\mathrm{lr}})$.

The whole WVTC transmission process is summarized as
\begin{equation}
	\begin{aligned}
		\mathbf V
		&\xrightarrow{\,f_{\mathrm{ds}}\,}
		\mathbf V^{\mathrm{lr}}
		\xrightarrow{\,f_{\mathrm{tok}}\,}
		\mathbf Z
		\xrightarrow{\,\Pi\,}
		\{\mathcal P_n\}_{n=1}^{N_p}
		\\[-1mm]
		&\quad
		\xrightarrow[\text{block fading transmission}]
		{\text{CSI-aware packet scheduling}}
		\widehat{\mathbf Z}
		\xrightarrow{\,f_{\mathrm{det}},\,f_{\mathrm{sr}}\,}
		\widehat{\mathbf V}.
	\end{aligned}
	\label{eq:end_to_end_process}
\end{equation}

\section{Importance- and CSI-Aware Loss-Resilient Token Transmission}

\begin{figure*}[htbp]
	\centering
	\includegraphics[width=6.6in]{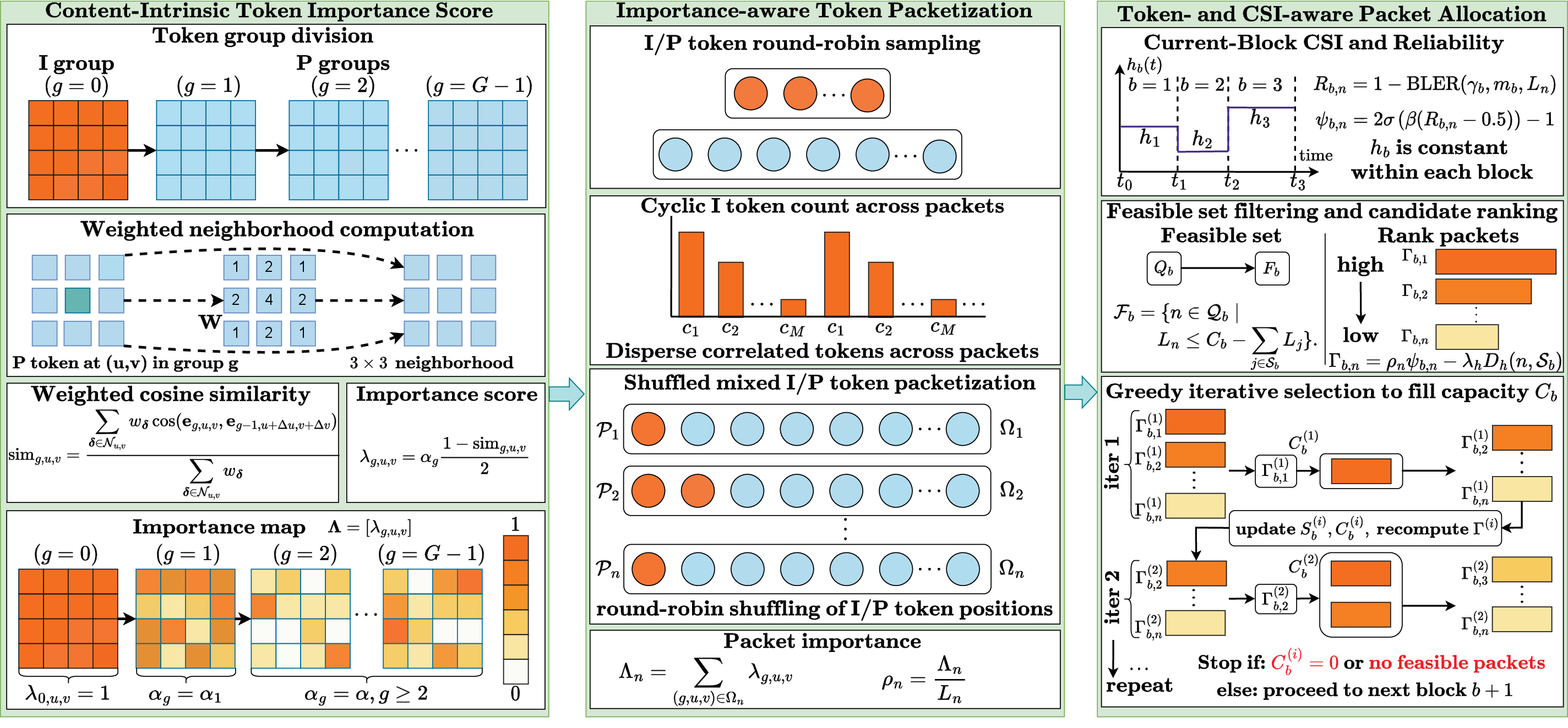}
	\vspace{-12pt}
	\caption{Illustration of the proposed importance- and CSI-aware
		loss-resilient token transmission.}
	\vspace{-12pt}
	\label{fig_2}
\end{figure*}

As shown in Fig.~\ref{fig_2}, the proposed scheme comprises
content-intrinsic token scoring, importance-aware packetization, and
online CSI-aware packet allocation. It disperses important and
correlated tokens across both packets and fading blocks.

\subsection{Content-Intrinsic Token Importance Score}

Let
$\Omega=\{(g,u,v)\mid 0\leq g<G,\;0\leq u<H_z,\;0\leq v<W_z\}$
denote all token positions, and
$\boldsymbol{\Lambda}\in\mathbb R^{G\times H_z\times W_z}$ the corresponding
importance map, where each entry $\lambda$ represents the importance score of the token at position $(g,u,v)$.

Before finite scalar quantization (FSQ), the tokenizer produces
$\mathbf E\in\mathbb R^{D\times G\times H_z\times W_z}$, which is
used only for importance evaluation, while its quantized indices form
$\mathbf Z$. Let
$\mathbf e_{g,u,v}=\mathbf E_{:,g,u,v}\in\mathbb R^D$.
Since I-tokens provide the structural anchor, we set
$\lambda_{0,u,v}=1$.

For P-tokens, importance is measured by temporal novelty relative to
the preceding token group. To tolerate local motion and spatial
misalignment, we adopt a convolution-inspired comparison over a
$3\times3$ neighborhood:
\begin{equation}
	\mathbf W=
	\begin{bmatrix}
		1 & 2 & 1\\
		2 & 4 & 2\\
		1 & 2 & 1
	\end{bmatrix}.
	\label{eq:spatial_weight}
\end{equation}

The $3\times3$ window captures nearby spatial correspondences with low
complexity. Moreover,
$\mathbf W=[1,2,1]^{\mathsf T}[1,2,1]$ forms a Gaussian-like
convolution kernel, assigning the largest weight to the co-located token
and smaller weights to surrounding neighbors.

Let $\boldsymbol{\delta}=(\Delta u,\Delta v)\in\{-1,0,1\}^2$ denote a
spatial offset,
$\mathcal N_{u,v}$ the valid neighborhood, and
$w_{\boldsymbol{\delta}}
=[\mathbf W]_{\Delta u+2,\Delta v+2}$ its weight. The weighted
similarity is
\begin{equation}
	\operatorname{sim}_{g,u,v}
	=
	\frac{
		\displaystyle
		\sum_{\boldsymbol{\delta}\in\mathcal N_{u,v}}
		w_{\boldsymbol{\delta}}
		\cos\!\left(
		\mathbf e_{g,u,v},
		\mathbf e_{g-1,u+\Delta u,v+\Delta v}
		\right)}
	{\displaystyle
		\sum_{\boldsymbol{\delta}\in\mathcal N_{u,v}}
		w_{\boldsymbol{\delta}}},
	\quad g\geq1.
	\label{eq:weighted_similarity}
\end{equation}

This operation resembles spatial convolution, but aggregates local
cosine similarities rather than feature values. For $g=1$, the
reference is the I-token group; otherwise, it is the preceding P-token
group.

The P-token importance is
\begin{equation}
	\lambda_{g,u,v}
	=
	\alpha_g
	\frac{1-\operatorname{sim}_{g,u,v}}{2},
	\qquad g\geq1,
	\label{eq:token_importance}
\end{equation}
where $\alpha_g=\alpha_1$ for the first P-token group and
$\alpha_g=\alpha$ otherwise. Lower similarity indicates greater
temporal novelty and thus higher importance.

Let $\Omega_n\subseteq\Omega$ denote the token positions in packet
$\mathcal P_n$. Its aggregate importance and importance density are
\begin{equation}
	\Lambda_n
	=
	\sum_{(g,u,v)\in\Omega_n}\lambda_{g,u,v},
	\qquad
	\rho_n
	=
	\frac{\Lambda_n}{L_n},
	\label{eq:packet_importance}
\end{equation}

Thus, $\rho_n$ measures importance per source bit and serves as the packet-level priority metric for the subsequent CSI-aware allocation.

\subsection{Importance-Aware Token Packetization}

Packing contiguous tokens together may cause clustered erasures, while
isolated I-token packets may lead to severe structural loss. We
therefore mix and disperse I- and P-tokens across packets.

Let $Q$ be the maximum number of token positions per packet. Packet
$\mathcal P_n$ contains
$a_n=c_{1+((n-1)\bmod M)}$ I-token positions, where
$\{c_1,\ldots,c_M\}$ is a predefined cycle of period $M$. The value
$a_n$ is adjusted near the end to assign every I-token exactly once
while satisfying the packet-size constraint.

Flattened I-token positions are traversed with a cyclic stride
$\kappa$, chosen to be coprime with the number of I-tokens. The
remaining $Q-a_n$ positions are filled with P-tokens permuted using a
shared pseudorandom seed and assigned in round-robin order.

Let $\Omega_n^{\rm I}$ and $\Omega_n^{\rm P}$ denote the I- and
P-token positions in packet $\mathcal P_n$. Then
$\Omega_n=\Omega_n^{\rm I}\cup\Omega_n^{\rm P}$ satisfies
\begin{equation}
	\Omega_n\cap\Omega_{n'}=\varnothing,\quad n\neq n',
	\qquad
	\bigcup_{n=1}^{N_p}\Omega_n=\Omega,
	\qquad
	|\Omega_n|\leq Q.
	\label{eq:token_partition}
\end{equation}

The packetization parameters, stride, and seed are shared by both
ends, allowing $\Omega_n$ to be recovered from the packet ID without
explicit position signaling. The resulting mixed packets disperse
structural anchors and correlated temporal regions while preserving
both contributions in their importance densities.

\subsection{Token- and CSI-Aware Packet Allocation}

Based on the packet importance density $\rho_n$, the retained packets
are allocated online using their importance and current channel state
information (CSI). For fading block $b$, let $\mathcal Q_b$ denote the
unscheduled packets. The scheduled set and remaining capacity are
initialized as $\mathcal S_b^{(0)}=\varnothing$ and
$C_b^{(0)}=C_b$, where $C_b^{(i)}$ is the remaining capacity after
$i$ selections.

At iteration $i$, the feasible packet set is
\begin{equation}
	\mathcal F_b^{(i)}
	=
	\left\{
	n\in\mathcal Q_b\setminus\mathcal S_b^{(i)}
	\;\middle|\;
	L_n\leq C_b^{(i)}
	\right\}.
	\label{eq:feasible_set}
\end{equation}

For each $n\in\mathcal F_b^{(i)}$, the decoding reliability $R_{b,n}$
is mapped to
\begin{equation}
	\psi_{b,n}
	=
	2\sigma\!\left(\beta(R_{b,n}-0.5)\right)-1,
	\label{eq:reliability_preference}
\end{equation}
where $\sigma(x)=1/(1+\exp(-x))$ and $\beta$ controls the transition
sharpness. Thus, $\rho_n\psi_{b,n}$ favors important packets with high
decoding reliability.

However, assigning too many important packets to the same block may
increase the risk of correlated erasures. We therefore define the
importance-concentration penalty as
\begin{equation}
	D_h\!\left(n,\mathcal S_b^{(i)}\right)
	=
	\sum_{j\in\mathcal S_b^{(i)}}\rho_n\rho_j.
	\label{eq:importance_concentration}
\end{equation}

A larger value indicates that the candidate packet would further
concentrate important information in the current block.

Combining reliability and concentration, the scheduling score is
\begin{equation}
	\Gamma_{b,n}^{(i)}
	=
	\rho_n\psi_{b,n}
	-
	\lambda_h
	D_h\!\left(n,\mathcal S_b^{(i)}\right),
	\label{eq:scheduling_score}
\end{equation}
where $\lambda_h\geq0$ controls the concentration penalty.

The highest-scoring feasible packet is selected, and the scheduled set
and remaining capacity are updated as
\begin{equation}
	\begin{aligned}
		n_i^\star
		&=
		\arg\max_{n\in\mathcal F_b^{(i)}}
		\Gamma_{b,n}^{(i)},\\
		\mathcal S_b^{(i+1)}
		&=
		\mathcal S_b^{(i)}\cup\{n_i^\star\},\\
		C_b^{(i+1)}
		&=
		C_b^{(i)}-L_{n_i^\star}.
	\end{aligned}
	\label{eq:greedy_update}
\end{equation}

The feasible set and scores are recomputed until no packet fits. The
final scheduled set is denoted by $\mathcal S_b$, and the remaining
packets are updated as
$\mathcal Q_{b+1}=\mathcal Q_b\setminus\mathcal S_b$.
Thus, the scheduler jointly considers packet importance, decoding
reliability, block capacity, and importance concentration using only
current block CSI.

\begin{figure*}[htbp]
	\centering
	\includegraphics[width=5.9in]{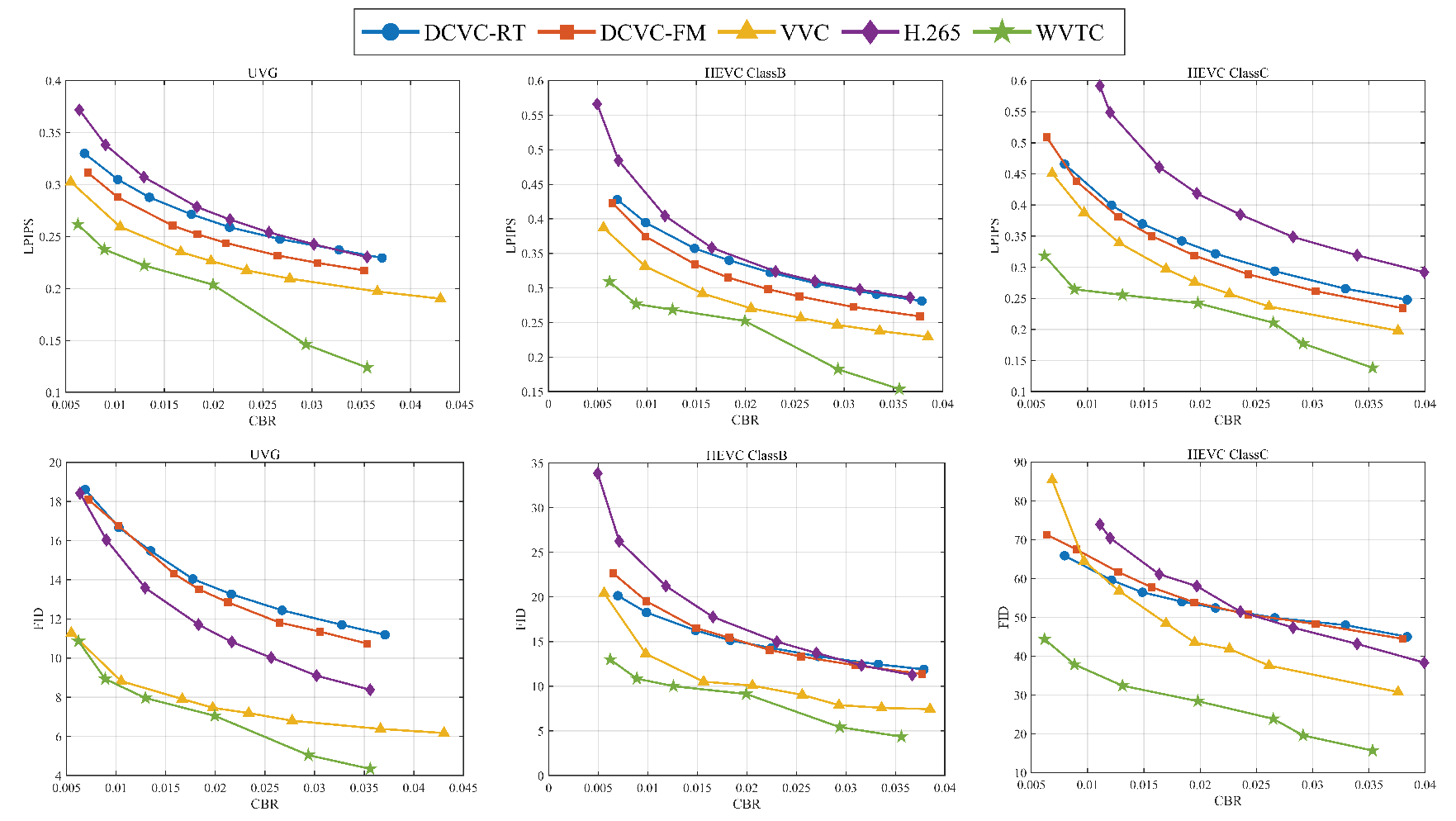}
	\vspace{-12pt}
	\caption{LPIPS and FID performance versus CBR on the UVG, HEVC ClassB, and HEVC ClassC sequences.}
	\vspace{-12pt}
	\label{fig_3}
\end{figure*}

\section{Deployment Details}

We fine-tune the receiver-side reconstruction network for robustness
against packet erasures. The pretrained VidTok \cite{vidtok} tokenizer and its FSQ codebook remain frozen, while the VidTok detokenizer and scale-specific SwinIR \cite{swinir} network are trainable. This
preserves the discrete token representation while adapting the receiver
to incomplete token tensors.

Training clips are sampled from Vimeo-90K \cite{vimeo}, randomly cropped to
$256\times256$, and divided into clips of $T=5$ frames. Each clip
$\mathbf X$ is downsampled, tokenized, and packetized using the proposed
shuffled mixed I/P-token mapping. To simulate structured transmission
losses, entire packets are randomly erased with a packet masking ratio
$r_m\sim\mathcal U(0,0.2)$, and all token positions carried by erased
packets are filled with zeros. The detokenizer reconstructs a
low-resolution video from the resulting incomplete token tensor, and
SwinIR produces the original-resolution frames.

The training objective combines the pixel-domain $\ell_1$ loss
$\mathcal L_{\rm pixel}$, learned perceptual image patch similarity (LPIPS) loss $\mathcal L_{\rm lpips}$, and
adversarial loss $\mathcal L_{\rm adv}$:
\begin{equation}
	\mathcal L
	=
	\mathcal L_{\rm pixel}
	+
	\mathcal L_{\rm lpips}
	+
	w_{\rm adv}\mathcal L_{\rm adv},
	\label{eq:training_loss}
\end{equation}
where $w_{\rm adv}$ controls the adversarial term.

For each retained packet $\mathcal P_n$, $n\in\mathcal Q_1$, its total
bit length is
\begin{equation}
	L_n
	=
	|\Omega_n|\left\lceil\log_2|\mathcal K|\right\rceil
	+
	\left\lceil\log_2N_p\right\rceil
	+
	B_{\rm CRC},
	\label{eq:packet_length}
\end{equation}
where the three terms correspond to the token payload, packet ID, and CRC bit overhead, respectively.

To measure the channel resources consumed by low-density parity check (LDPC) coding and quadrature amplitude
modulation (QAM), we define the channel bandwidth ratio (CBR)~\cite{djscc} as
\begin{equation}
	\mathrm{CBR}
	=
	\frac{
		\displaystyle
		\sum_{b=1}^{B}
		\sum_{n\in\mathcal S_b}
		\left\lceil
		\frac{L_n}
		{r(m_b)\log_2\mathcal M(m_b)}
		\right\rceil}
	{T H_0W_0},
	\label{eq:cbr}
\end{equation}
where $r(m_b)$ and $\mathcal M(m_b)$ denote the channel coding rate and
modulation order of MCS $m_b$, respectively. During inference, the CBR
is adjusted through the spatial downsampling factor $\times[2,3,4]$, clip length $T$,
and the exclusion of low-importance-density packets when constructing
$\mathcal Q_1$.

\section{Numerical Results}
In this section, we present numerical results to evaluate the effectiveness of proposed WVTC.

\subsection{Experimental Setups}
\subsubsection{Datasets}

We evaluate WVTC on the UVG and the HEVC dataset, using the first 100 frames of each video.

\subsubsection{Model Deployment Details}
We employ the discrete FSQ variant of VidTok with codebook cardinality
$|\mathcal K|=4096$ and fixed $Q=390$ token positions and $B_{\rm CRC}=24$ bits. The channel follows independent block Rayleigh fading. Unless otherwise specified, we use the varying-SNR mode, where the instantaneous block SNR is \(\gamma_b=\bar{\gamma}|h_b|^2\), with average \({\gamma}_b=8\) dB. The MCS Table is constructed based on MCS Table I of 3GPP TS~38.214 \cite{sionna}, and BLER is estimated from the offline Rayleigh BLER lookup table. We use the same Rayleigh seed, channel realizations, and MCS-selection policy for all schemes.

For fine-tuning, the per-process batch size is 4. The optimizer is AdamW,
with learning rate \(5\times10^{-6}\) for both the generator-side network and
the adversarial discriminator. We set \(\alpha_1=0.9\) for the computed first P-token importance,
\(\alpha=0.75\), \(M=5\), \(\beta=2\), \(\kappa=32\),
\(\lambda_h=2\), and \(w_{\rm adv}=0.02\). All experiments are conducted on NVIDIA GeForce RTX 5090 GPUs using PyTorch 2.11.0.

\subsubsection{Comparison Benchmarks}
In the experiments, we consider the following benchmarks.

$\textbf{DCVC-RT}$: A real-time deep contextual video compression (DCVC) framework with compact coding structures \cite{dcvc-rt}.

$\textbf{DCVC-FM}$: A feature-modulated DCVC framework that supports rate adaptation over a wide range
of bitrates \cite{dcvc-fm}.

$\textbf{VVC}$: The traditional Versatile Video Coding (VVC) standard
\cite{vvc}.

$\textbf{H.265}$: The traditional High Efficiency Video Coding (H.265)
standard.

For a fair comparison, all schemes employ the same 5G MCS configuration, with LDPC coding and QAM implemented using Sionna \cite{sionna}. VVC and H.265 are implemented using FFmpeg 8.1. For the CBR comparison, every channel use consumed by retransmission is included in the CBR of the benchmark schemes. WVTC does not invoke retransmission and reconstructs the video directly from the successfully received token packets.

\subsubsection{Evaluation Metrics}

We employ LPIPS and Fr\'echet Inception Distance (FID) to evaluate perceptual quality.

\subsection{Results Analysis}

\subsubsection{Performance for Different CBRs}

Fig.~\ref{fig_3} compares the LPIPS and FID performance of the considered schemes across different CBRs. As the CBR increases, both LPIPS and FID generally decrease because more source and channel
resources are available for reconstruction. WVTC achieves the lowest
or near-lowest LPIPS and FID across most operating points, indicating
better perceptual quality and distributional similarity to the original
frames. At low CBRs, all schemes are constrained by the limited amount
of transmitted information, whereas the advantage of WVTC becomes more
pronounced in the medium- and high-CBR regions. This is because more
informative token packets can be retained and delivered, while the
importance-aware packetization disperses critical structural and
temporal information and the CSI-aware scheduler avoids concentrating
important packets in unreliable fading blocks. The consistent gains
across sequences with different spatial content and motion
characteristics further demonstrate the effectiveness and robustness of
the proposed packetization and allocation strategy.

\subsubsection{Performance for Different PERs}

\begin{figure}[htbp]
	\centering
	\includegraphics[width=3.0in]{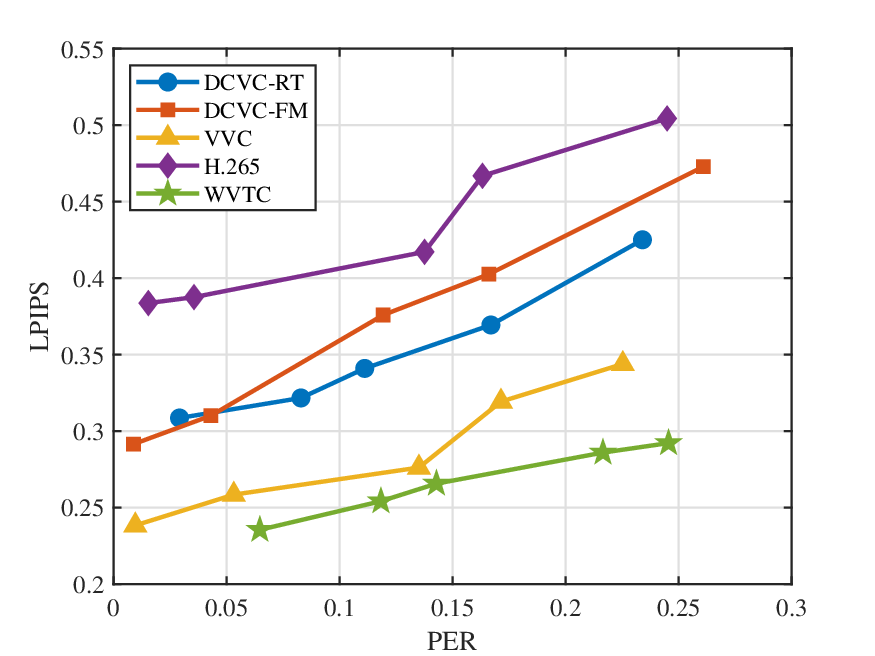}
	\vspace{-12pt}
	\caption{LPIPS performance under different PERs.}
	\vspace{-6pt}
	\label{fig_4}
\end{figure}

Fig.~\ref{fig_4} compares the perceptual reconstruction quality under different packet error rate (PERs) on HEVC ClassC. As the PER increases, all schemes exhibit performance degradation because more coded information is unavailable at the receiver. Nevertheless, WVTC exhibits more graceful perceptual degradation and remains the best-performing scheme over most of the evaluated range. Conventional and learned predictive codecs are sensitive to the loss of reference information and suffer increasingly severe error propagation. In contrast, WVTC disperses structural and temporally correlated tokens across packets, allocates important packets according to the current block reliability, and reconstructs the video from the surviving token context. These results demonstrate that the proposed design provides superior loss resilience under unreliable block fading transmission.

\section{Conclusion}

This paper proposed a loss-resilient wireless video token communication framework for block fading channels. WVTC combines content-intrinsic token-importance evaluation, shuffled mixed I/P-token
packetization, and online CSI-aware packet allocation to disperse
important and temporally correlated information across packets and
fading blocks. At the receiver, a fine-tuned detokenizer reconstructs
missing content from the surviving token context without retransmission.
Numerical results show that WVTC achieves improved perceptual quality across different CBRs and exhibits more graceful degradation as the packet error rate increases. These results demonstrate the effectiveness of jointly exploiting video-token structure and current channel conditions for
robust wireless video transmission.

\end{document}